%% file: ms.tex
\documentclass{bmvc2k}
\usepackage{amssymb}
\usepackage{tabularx}

\title{Propagating Confidences through CNNs for Sparse Data Regression}

\addauthor{Abdelrahman Eldesokey}{abdelrahman.eldesokey@liu.se}{1}
\addauthor{Michael Felsberg}{michael.felsberg@liu.se}{1}
\addauthor{Fahad Shahbaz Khan\textsuperscript{1,}}{fahad.khan@liu.se}{2}

\addinstitution{
 Computer Vision Laboratory\\
 Department of Electrical Engineering \\
 Link\"oping University\\
 Link\"oping, Sweden
}
\addinstitution{
Inception Institute of Artificial Intelligence\\
Abu Dhabi, UAE
}

\runninghead{Eldesokey \etal}{Propagating Confidences through Regression CNNs}

\def\eg{\emph{e.g}\bmvaOneDot}

\def\etal{\emph{et al}\bmvaOneDot}

\def\C{\mathbb{C}}
\newcolumntype{L}[1]{>{\raggedright\arraybackslash}p{#1}}
\newcolumntype{C}[1]{>{\centering\arraybackslash}p{#1}}
\newcolumntype{R}[1]{>{\raggedleft\arraybackslash}p{#1}}
\begin{document}

\maketitle

\begin{abstract}
In most computer vision applications, convolutional neural networks (CNNs) operate on dense image data generated by ordinary cameras. Designing CNNs for sparse and irregularly spaced input data is still an open problem with numerous applications in autonomous driving, robotics, and surveillance. To tackle this challenging problem, we introduce an algebraically-constrained convolution layer for CNNs with sparse input and demonstrate its capabilities for the scene depth completion task. We propose novel strategies for determining the confidence from the convolution operation and propagating it to consecutive layers. Furthermore, we propose an objective function that simultaneously minimizes the data error while maximizing the output confidence. Comprehensive experiments are performed on the KITTI depth benchmark and the results clearly demonstrate that the proposed approach achieves superior performance while requiring three times fewer parameters than the state-of-the-art methods. Moreover, our approach produces a continuous pixel-wise confidence map enabling information fusion, state inference, and decision support.

\end{abstract}

\section{Introduction}
\label{sec:intro}
\input{sec/intro}

\section{Related Work}
\label{sec:related}
\input{sec/related}

\section{Our Approach}
\label{sec:method}
\input{sec/method}

\section{Experiments}
\label{sec:exp}
\input{sec/exp}
\vspace{-5px}
\section{Conclusion}
\label{sec:concl}
\input{sec/concl}

\section{Acknowledgments}
This research is funded by Vinnova through grant CYCLA, the Swedish Research Council through a framework grant for the project Energy Minimization for Computational Cameras (2014-6227), CENIIT grant (18.14) and VR starting grant (2016-05543).

\newpage

\bibliography{refs}

\end{document}

%% file: sec/intro.tex
In recent years, machine learning methods have achieved significant successes in many computer vision applications, making use of data from monocular passive image sensors, such as grayscale, RGB, and thermal cameras. Typically, data generated by these image sensors are dense and most existing machine learning methods are designed to fully exploit this dense data in order to understand the scene content. Different to the aforementioned sensors, active sensors, such as LiDAR, RGB-D, and ToF cameras, produce sparse data. Here, the sparse output is caused by the acquisition process through active sensing compared to passively measuring light influx in conventional 2D sensors with dense output. The sparse output imposes additional challenges on the machine learning methods to infer the missing data and find an accurate reconstruction of the entire scene.  

\begin{figure}[t]
\begin{tabular}{cccc}
\includegraphics[width=0.24\textwidth]{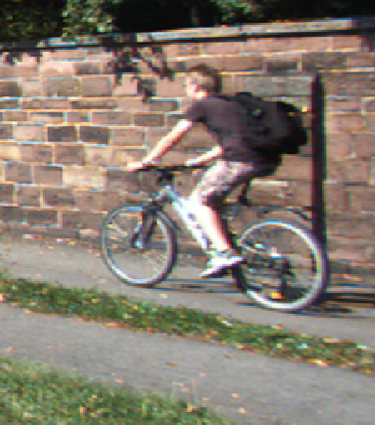}&
\includegraphics[width=0.24\textwidth]{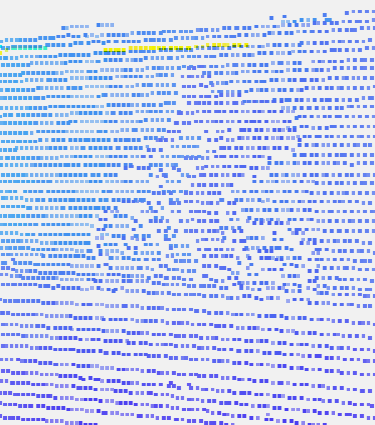}&
\includegraphics[width=0.24\textwidth]{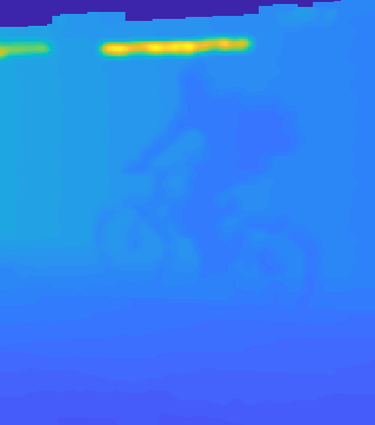}&
\includegraphics[width=0.24\textwidth]{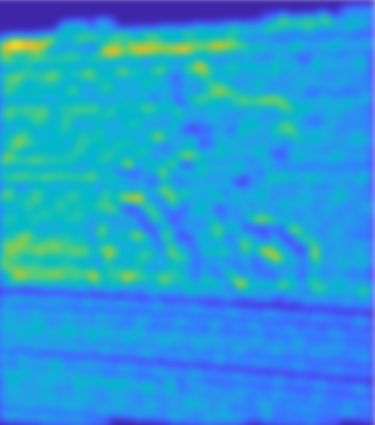} \\
(a) RGB image & (b) LiDAR data* & (c) Dense output & (d) Output confidence
\end{tabular}
\caption{Depth map completion example. The RGB image (a) associated with a projected LiDAR point cloud in (b).  The depth map completion output is shown in (c) together with pixel-wise confidence (d) (blue is low and yellow is high). Most existing deep learning methods struggle in scenarios as shown here due to the very high sparsity of the input data (95\% of pixels are missing). [*The image is dilated for the sake of visibility]}
\label{fig:intro}
\end{figure}

Sensors with sparse outputs are becoming increasingly popular and have numerous applications in autonomous driving, robotics, and surveillance due to their range measuring capability. One fundamental task is \emph{scene depth completion} that aims to reconstruct a full depth map from sparse input. Scene depth completion is a required processing step in, \eg situation awareness and decision support. One of the key challenges when tackling the problem of scene depth completion is the handling of missing values while also differentiating them from the zero-valued regions. Besides, densifying the depth map, corresponding confidences are also desirable since they provide information about reliability of the output values. Such confidence maps are highly important for decision making in safety applications, \eg obstacles detection in autonomous vehicles and robotics. Figure \ref{fig:intro} shows an example of a depth completion task. Given the projected LiDAR point cloud, the objective is to densify the sparse depth map, either utilizing the RGB image (guided completion) or only using the projected point cloud (unguided completion). The output is a complete dense map together with pixel-wise output confidence. 

\setlength{\tabcolsep}{1pt}

Recently, deep learning, notably Convolutional Neural Networks (CNNs), have demonstrated great potential in solving a variety of computer vision tasks. Generally, CNNs are formed by several convolution, local normalization, and pooling layers. The final layers of CNNs are often fully connected (FC) where for classification problems, the last FC layer employs a softmax function to approximate the probability or confidence over the class memberships. Such confidences are often missing in regression settings, although both the regressed value \emph{and} its confidence are required for numerous applications. For example, it is not only relevant to know how far away a potential obstacle is located, but also how reliable this information is. For the scene depth completion task, several deep regression networks have been proposed in the literature that introduce confidence measures \cite{ren2015shepard, Uhrig2017, Chodosh2018, Liu2018}. However, all these methods utilize the confidence as a binary valued mask to filter out the missing measurements. This strategy disregards valuable information available in the confidence maps. Different to these existing methods, we propose an approach that utilizes the signal confidence as a continuous measure for data uncertainty and propagates it through all layers.

In this paper, we propose an algebraically-constrained convolution operator for deep networks with sparse input to achieve a proper processing of confidences. The sparse input is equipped with confidences and the network is required to produce a dense output. We derive novel methods for determining the confidence from the convolution operation and propagating it to consecutive layers. To maintain the confidences within a valid range, we impose non-negativity constraints on the network weights during training. Further, we also introduce an objective function that simultaneously minimizes the data error while maximizing the output confidence. Moreover, we demonstrate the significance of the proposed confidence measure by introducing a novel approach for performing scale-fusion based on confidences. Our proposed method achieves state-of-the-art results on the KITTI depth benchmark \cite{Uhrig2017} while requiring only 480 parameters, which is three times fewer than state-of-the-art methods.

%% file: sec/related.tex
Scene depth completion is a challenging problem that aims to construct a dense depth map given a sparse depth image. It shares similarities with image inpainting since both tasks require filling missing information/pixels in an image. In case of image inpainting, several approaches based on deep learning have been introduced recently, however restricted to binary masks. These masks define regions in the image where missing pixels have zero values and the remaining pixels have ones. K{\"o}hler \etal \cite{kohler2014mask} showed quantitatively how incorporating those binary masks in training shallow networks leads to better results, even if the masks were not available during test time. Ren \etal \cite{ren2015shepard} proposed a convolution operation based on Shepard interpolation \cite{shepard1968two} that also utilizes a binary mask to perform inpainting or super-resolution. They propagated the binary masks by convolving them with the same filters/weights as the data and thresholding insignificant values. Liu \etal \cite{Liu2018} incorporated the use of binary masks in the U-Net architecture \cite{Ronneberger2015} for performing inpainting. The binary masks were propagated by setting the pixel at the filter origin to one if not all pixels within the filter support are unknown.

In case of scene depth completion, Uhrig \etal introduced the KITTI depth benchmark \cite{Uhrig2017} which is a large-scale dataset for this task. They also proposed a method, where the convolution operations are weighted using the binary masks. The masks are propagated using the max pooling operation. In their work, they also investigated concatenating the binary mask to the input as an additional channel. Chodosh \etal \cite{Chodosh2018} utilized compressed sensing to approach the sparsity problem for scene depth completion. A binary mask is employed to filter out the unmeasured values. Further, their method requires significantly fewer parameters compared to \cite{Uhrig2017}. Ma and Karaman \cite{Ma2017SparseToDense} proposed sparse-to-dense deep network which utilizes an RGB image and a randomly sampled set of sparse depth measurements to produce a dense depth map.

Our approach is different to the aforementioned methods in several aspects. Firstly, we treat the binary masks as continuous confidences instead of binary values. We further derive an algebraically-constrained deep convolution operator from the normalized convolution framework \cite{Knutsson} that infers continuous output confidences. Secondly, different to \cite{ren2015shepard}, we enforce the trained filters to be positive to obtain a sound confidence. This is also different from \cite{Uhrig2017, Liu2018} that employed a constant averaging filter for the confidences, a strategy that assumes a uniform confidence distribution among all pixels, which is generally not the case in real-world data. Thirdly, we do not constrain the output confidences to be binary as in \cite{ren2015shepard, Uhrig2017, Liu2018}. Instead, we propose a computational scheme that allows output confidences to be continuous \emph{while} propagating confidence information from the input to the output. Finally, we demonstrate that utilizing normalized convolution to perform scale-fusion in multi-scale networks based on confidences outperforms the standard convolution used in, \eg in U-Net \cite{Ronneberger2015, Liu2018}. Moreover, our proposed approach requires remarkably fewer parameters compared to the aforementioned approaches, while achieving state-of-the-art results.

%% file: sec/method.tex
Here, we describe our approach by starting with a brief introduction to the normalized convolution framework. We then introduce an algebraically-constrained  normalized convolution operator for CNNs and the propagation method for confidences. Finally, we describe our proposed network architecture and the loss function taking confidences into account.

\subsection{Normalized Convolution}
Assume a sparse signal/image $\mathbf{F}$ with missing parts due to noise, acquisition process, preprocessing, or other system deficiencies. The missing parts of the signal are identified using a confidence mask $\mathbf{C}$, which has zeros or low values at missing/uncertain locations and ones otherwise. The signal is sampled and, at each sample point $k$, the neighborhood is represented as a finite dimensional vector $\mathbf{f}_k \in \mathbb{C}^n$ accompanied with a confidence vector $\mathbf{c}_k$ of the same size, both assumed to be column vectors. Using the notation from \cite{farneback:phd_thesis}, \emph{normalized convolution} is defined at all locations $k$ as (index $k$ is omitted to reduce clutter):
\begin{equation}\label{eq:1}
\mathbf{r} = (\mathbf{B}^* \mathbf{D}_\mathbf{a} \mathbf{D}_\mathbf{c} \mathbf{B})^{-1} \mathbf{B}^* \mathbf{D}_\mathbf{a} \mathbf{D}_\mathbf{c} \mathbf{f}\enspace,
\end{equation}
where $\mathbf{B}$ is a matrix which incorporates a set of basis functions $ \{ \mathbf{b}_i \in \C^n \}_1^m $ in its columns, $ \mathbf{D}_\times $ denotes a diagonal matrix with vectorized $\times$ on the diagonal, $\mathbf{a}$ is the applicability function which is a \emph{non-negative} localization function for the basis $\mathbf{B}$, and $\mathbf{r}$ holds the coefficients of the signal $\mathbf{F}$ at location $k$ projected onto the subspace spanned by $\mathbf{B}$.

The basis functions in $\mathbf{B}$ could be polynomials or complex exponentials, but the simplest case is when $\mathbf{B} = \{ \mathbf{b}_\mathbf{1}:\mathbf{b}_\mathbf{1} = \mathbf{1}\} $, and it becomes \emph{normalized averaging}. 
In this case, the signal is mapped onto a constant, localized with the applicability function, and (\ref{eq:1}) simplifies to 
$\mathbf{r} = (\mathbf{1}^* \mathbf{D}_\mathbf{a} \mathbf{D}_\mathbf{c} \mathbf{1})^{-1} \mathbf{1}^* \mathbf{D}_\mathbf{a} \mathbf{D}_\mathbf{c} \mathbf{f}$,
which can be formulated for the full signal $\mathbf{F}$ and its confidence $\mathbf{C}$ as ($\ast$ denotes convolution and $\cdot$ point-wise multiplication):
\begin{equation} \label{eq:3}
	\mathbf{r}[k] = \dfrac{\sum_i \ \mathbf{a}[i] \  \mathbf{F}[k-i] \ \mathbf{C}[k-i]}{\sum_i \  \mathbf{a}[i] \ \mathbf{C}[k-i]} = \dfrac{\mathbf{a} \ast (\mathbf{F} \cdot \mathbf{C}) }{\mathbf{a} \ast \mathbf{C}} [k]\enspace,
\end{equation}


\subsection{Training the Applicability}
The appropriate choice of the applicability function is an open issue as it usually depends on the nature of the data. Therefore, methods for statically estimating the applicability function have been suggested 
\cite{MM_SNA}, but we aim to learn $\mathbf{a}$ as part of the 
training. This generalizes convolutional layers, as normalized averaging is equivalent to standard convolution in case of signals with constant confidence.  As described above, the applicability function acts as a confidence or localization function for the basis and therefore it is essentially non-negative. 

Non-negative  applicabilities are feasible to train in standard frameworks,  since back-propagation is based on the chain rule and any differentiable function with non-negative co-domain can be plugged in. Thus, the function $\Gamma(\cdot)$, e.g. the softplus, is applied to the  the weights $\mathbf{W}$, and the gradients for the weight element $\mathbf{W}^l_{m,n}$ at the $l^\mathrm{th}$ convolution layer are calculated as:
\begin{equation}\label{eq:4}
\frac{\partial \mathbf{E}}{\partial \mathbf{W}^l_{m,n}} = \sum_{i,j} \frac{\partial \mathbf{E}}{\partial \mathbf{Z}^l_{i,j}} . \frac{\partial \mathbf{Z}^l_{i,j}}{\partial \ \Gamma (\mathbf{W}^l_{m,n})} . \frac{\partial \ \Gamma (\mathbf{W}^l_{m,n})}{\partial \mathbf{W}^l_{m,n}}\enspace,
\end{equation}
where $\mathbf{E}$ is the loss between the output and the ground truth,  $\mathbf{Z}^l_{i,j}$ is the output of the $\emph{l}^\mathrm{th}$ layer at locations $i,j$ that were convolved with the weight element $\mathbf{W}^l_{m,n}$. Accordingly, the forward pass for \emph{normalized convolution} is defined as:
\begin{equation}\label{eq:5}
\mathbf{Z}^l_{i,j} =  \frac{\sum_{m,n} \ \mathbf{Z}^{l-1}_{i+m,j+n} \mathbf{C}^{l-1}_{i+m,j+n} \ \Gamma (\mathbf{W}^l_{m,n})}{ \sum_{m,n} \ \mathbf{C}^{l-1}_{i+m,j+n} \ \Gamma(\mathbf{W}^l_{m,n}) + \epsilon}\enspace,
\end{equation}
where $\mathbf{C}^{l-1}$ is the confidence from the previous layer, $\mathbf{W}^l_{m,n}$ is the applicability in this context and $\epsilon$ is a constant to prevent division by zero. Note that this is formally a correlation, as it is a common notation in CNNs.


\subsection{Propagating Confidence}
The main strength about the signal/confidence philosophy is the availability of confidence information apart from the signal.  This is needed to be propagated through the network to output a pixel-wise confidence aside with the network prediction. In normalized convolution frameworks, Westelius \cite{Westelius302463} proposed a measure for propagating certainties:
\begin{equation}\label{eq:6}
\mathbf{C}_\mathrm{out} = \Big( \frac{\det \ \mathbf{G} }{\det \ \mathbf{G}_0 } \Big)^{\frac{1}{m}}\enspace,
\end{equation}
where $\mathbf{G} = (\mathbf{B}^* \mathbf{D}_a \mathbf{D}_c \mathbf{B})$, $\mathbf{G}_0 = (\mathbf{B}^* \mathbf{D}_a \mathbf{B})$ and  $m$ is the number of basis functions. 

This measure calculates a geometric ratio between the Grammian matrix $\mathbf{G}$ in case of partial confidence and $\mathbf{G}_0$ in case of full confidence. Setting $\mathbf{B} = \{ \mathbf{b}_\mathbf{1}:\mathbf{b}_\mathbf{1} = \mathbf{1}\} $, i.e., $m=1$,
we can utilize the already-computed term in (\ref{eq:5}) to propagate the confidence as follows:
\begin{equation}
\mathbf{C}^l_{i,j} = \frac{ \sum_{m,n} \mathbf{C}^{l-1}_{i+m,j+n} \ \Gamma( \mathbf{W}^l_{m,n}) + \epsilon}{\sum_{m,n} \ \Gamma(\mathbf{W}^l_{m,n})}\enspace,
\end{equation}


\subsection{Loss Function}
For scene depth completion task, we usually aim to minimize a norm, \textit{e.g}. \emph{l1} or \emph{l2} norm, between the output from the network and the ground truth. In our proposed method, we use the Huber norm, which is a hybrid between the \emph{l1} and the \emph{l2} norm and it is defined as:
\begin{equation}
\|z- t\|_\mathrm{H} =
\begin{cases} 
0.5(z-t)^2 & |z-t| < 1 \\ 
|z-t|-0.5, & \text{otherwise}
\end{cases}
\end{equation}
The Huber norm helps preventing exploding gradients in case of highly sparse data, which stabilizes the convergence of the network.
Nonetheless, our aim is not only to minimize the error norm between the output and the groundtruth, but also to increase the confidence of the output data. Thus, we propose a new loss which has a data term \emph{and} a confidence term:
\begin{equation}
 \mathbf{E}_{i,j} = \|\mathbf{Z}^L_{i,j} - \mathbf{T}_{i,j} \|_\mathrm{H} \enspace, \qquad \tilde{\mathbf{E}}_{i,j} = \mathbf{E}_{i,j} - \frac{1}{p} \left[ \mathbf{C}^L_{i,j} - \mathbf{E}_{i,j} \mathbf{C}^L_{i,j} \right] \enspace,
\end{equation}
where $\mathbf{Z}^L_{i,j}$ is the data output from the final layer $L$, $\mathbf{C}^L_{i,j}$ is the corresponding confidence output, $\mathbf{T}_{i,j}$ is the ground truth, $p$ is the epoch number, and $\|\cdot\|_\mathrm{H}$ is the Huber norm. The main objective of the loss is to minimize the error of the data term and maximize the confidence of the output. Note that the third term in the loss prevents the confidence from growing indefinitely. We weight the confidence term using the reciprocal of the epoch number $p$ to prevent it from dominating the loss function when the data error starts to converge.

\begin{figure}[tbp]
\centering
\includegraphics[width=\textwidth]{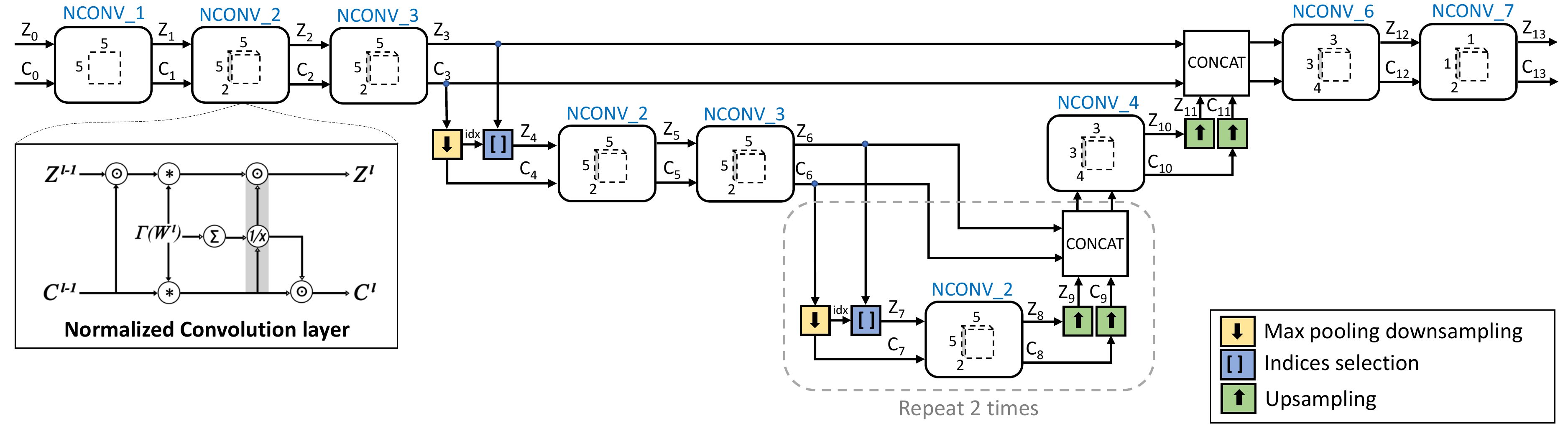}
\caption{Our proposed multi-scale architecture for the task of scene depth completion which utilizes normalized convolution layers. Downsampling is performed using max pooling on confidence maps and the indices of the pooled pixels are used to select the same pixels from the feature maps. Different scales are fused by upsampling the coarser scale and concatenate it with the finer scale. A normalized convolution layer is then used fuse the feature maps based on the confidence information. Finally, a 1 $\times$ 1 normalized convolution layer is used to merge different channels into one channel dense output and output confidence map. }
\label{fig:arch}
\end{figure}

\subsection{Network Architecture}
Inspired by \cite{Ronneberger2015}, we propose a hierarchical multi-scale architecture that shares the same weights between different scales, which leads to a very compact network as shown in Figure \ref{fig:arch}. Downsampling is performed using max pooling on the confidences and similar to \cite{zeiler2014visualizing} we keep the indices of the pooled pixels, which are then used to select the same pixels from the feature maps, \emph{i.e.}, we keep the most confident feature map pixels. The downsampled confidences are divided by the Jacobian of the scaling to maintain absolute confidence levels. Scale fusion is performed by upsampling the coarser scale and concatenate it with the finer scale. We apply a normalized convolution operator on the concatenated feature map to allow the network to fuse different scales utilizing confidence information. 



%% file: sec/exp.tex
\subsection{Experimental Setup}

\noindent\textbf{Dataset}: We evaluate our method on the KITTI depth benchmark \cite{Uhrig2017} which consists of projected LiDAR point clouds. The resulting depth maps/images are very sparse (approximately 4\% of pixels have values). The benchmark has 86,000 training images, 7,000 validation images and 1,000 test server images with no access to the ground truth. The ground truth has missing parts as it was matched with the stereo disparity to remove projected LiDAR outliers. We evaluate on the \emph{full} validation set as in \cite{Chodosh2018} and the test set.

\noindent\textbf{Implementation details}: All our experiments are performed on a workstation with Intel Xeon CPU (4 cores), 8 GB of RAM and NVIDIA GTX 1080 GPU with 8 GB of memory. NConv-HMS, NConv-1Scale(4ch), and NConv-SF-STD are trained with a batch size of $8$, while NCONV-1-Scale(16ch) are trained with a batch size of $4$. Our networks were trained on the first 10,000 out of 86,000 depth maps/images in the training set. We use the ADAM solver with default parameters except for the learning rate which we set to 0.01. \footnote{The source code will be made publicly available.}

\noindent\textbf{Evaluation metrics}: For comparison, we use the same evaluation metrics as defined in \cite{Chodosh2018, Ma2017SparseToDense}: \emph{Mean Absolute Error} (MAE) which is an unbiased error metric, \emph{Root Mean Square Error} (RMSE) which penalizes large errors, \emph{Mean Absolute Relative Error} (MRE) is a ratio between the error magnitude and the groundtruth value, and \emph{Inliers Ratio} ($\delta_i$) which is the percentage of pixels having relative error less than a specific threshold to the power of $i$. As in \cite{Chodosh2018}, we use a challenging threshold value of $\delta=1.01$.


\begin{table}
\begin{center}
\begin{tabular}{R{3.5cm} | C{1cm} C{1cm} C{1cm} C{1cm} C{1cm} C{1cm} C{1.5cm} C{1cm}}
 & MAE [m] & RMSE [m] & MRE  & $\delta<1.01$ & $\delta<1.01^2$ & $\delta<1.01^3$ &  \#Params & Output Conf.\\
\hline 
CNN \cite{Uhrig2017} & 0.78 & 2.97 & - & - & - & - & $2.5 \times 10^4$ & No \\
CNN+mask \cite{Uhrig2017} &  0.79 & 2.24 & - & - & - & - & $2.5 \times 10^4$ & No \\
SparseConv \cite{Uhrig2017} & 0.58 & 1.80  & 0.035 & 0.33  & 0.65  & 0.82  & $2.5 \times 10^4$ & No \\
Sparse-To-Dense \cite{Ma2017SparseToDense} & 0.70 & 1.68 & 0.039 & 0.21 & 0.41 & 0.59 & $3.4 \times 10^6$ & No \\
DCCS-1-Layer \cite{Chodosh2018} & 0.83 & 2.77 & 0.054 & 0.30 & 0.47 & 0.59 & $1.0 \times 10^3$ & No \\
DCCS-2-Layers \cite{Chodosh2018} & 0.47 & 1.45 & 0.028 & 0.41 & 0.68 & 0.80 & $1.8 \times 10^3$ & No \\
DCCS-3-Layers \cite{Chodosh2018} & 0.43 & \textbf{1.35} & 0.024 & 0.48 & 0.73 & 0.83 & $1.7 \times 10^3$ & No \\
\hline
NConv-1-Scale(16ch) & 0.40  & 1.58 & 0.022 &\textbf{ 0.60} & \textbf{0.81} & 0.88 & $2.5 \times 10^4$ & Yes \\ 
NConv-1-Scale(4ch) & 0.42 & 1.59 & 0.022 & 0.59 & 0.80 & 0.88 & $2.0 \times 10^3$ & Yes \\ 
NConv-HMS & \textbf{0.38} & {1.37} &\textbf{ 0.021} & \textbf{0.60} & \textbf{0.81} & \textbf{0.89} & $\mathbf{4.8 \times 10^2}$ & Yes \\ 
NConv-SF-STD & {0.53} & {3.0} &{0.037} & {0.59} & {0.80} & {0.88} & $\mathbf{4.8 \times 10^2}$ & No \\ 
\hline
\end{tabular}
\end{center}
\caption{Evaluation results on the validation set. The results for CNN and CNN+mask are taken from \cite{Uhrig2017}, SparseConv, Sparse-To-Dense and DCCS are from \cite{Chodosh2018}. Our multi-scale architecture NConv-HMS outperforms all other method in all evaluation metrics except for RMSE, where it is slightly inferior to DCCS-3-Layers.}
\label{tab:1}
\end{table}

\subsection{Quantitative Comparisons}
We compare our method with state-of-the-art methods in the literature: Sparsity Invariant Convolution (SparseConv) \cite{Uhrig2017}, Deep Convolutional Compressed Sensing (DCCS) \cite{Chodosh2018}, and Sparse-To-Dense \cite{Ma2017SparseToDense} approaches. As mentioned earlier, the Sparsity Invariant Convolution method applies a constrained convolution operation using binary masks. The DCCS approach \cite{Chodosh2018} employs compressed sensing and Alternating Direction Neural Networks (ADNNs) to create a deep auto-encoder that constructs a dense output. The Sparse-To-Dense method \cite{Ma2017SparseToDense} utilizes a ResNet architecture to encode the sparse LiDAR point clouds and RGB images and then decode a dense output

\noindent\textbf{Impact of continuous confidences}: 
To evaluate the impact of employing our proposed confidence scheme, we evaluate a single-scale architecture as described in \cite{Uhrig2017}. This architecture consists of 6 normalized convolution layers with filter sizes of $11\times11, \ 7\times7, \ 5\times5, \ 3\times3, \ 3\times3 \ \text{and} \ 1\times1 $ respectively with \emph{16 channels} each and we denote it as {NConv-1-Scale(16ch)}. To further demonstrate the efficiency of our approach, we evaluate the same architecture with 4 channels only and we denote it as NConv-1-Scale(4ch). Table \ref{tab:1} shows the results for both experiments as well as other methods in comparison. Our single-scale architecture {NConv-1-Scale(16ch)} achieves superior results in terms of MAE, MRE and $\delta_i$ compared to all other methods. This demonstrates the advantage of our proposed confidences scheme compared to SparseConv \cite{Uhrig2017}. Moreover, our compact architecture {NConv-1-Scale(4ch)} maintains the performance while requiring remarkably fewer parameters. However, DCCS-2-Layers and DCCS-3-Layers achieve better RMSE than our proposed single-scale architecture, which we attribute to the insufficient receptive field of the network.

\noindent\textbf{Multi-scale architecture}: 
To address the problem of the limited receptive field of our single-scale architecture, we incorporate a multi-scale architecture inspired by \cite{Ronneberger2015}. We further maintain the low number of parameters by sharing the weights/filters between different scales. The multi-scale architecture is illustrated in Figure \ref{fig:arch} and denoted as {NConv-HMS}. Table \ref{tab:1} provides the comparison between {NConv-HMS} and existing methods. Our NConv-HMS achieves better results compared to the single-scale architectures with respect to all the evaluation metrics. The RMSE is the most significantly reduced measure and becomes almost the same as for DCCS-3-Layers. Note also that the number of parameters was reduced to 480, which is remarkably fewer than all other methods in comparison.

\noindent\textbf{Impact of proposed scale-fusion scheme}: 
A common approach to perform multi-scale fusion is to upsample the coarser scales, concatenate it with the finer scale and then use a convolution layer to learn the proper fusion as in \cite{Ronneberger2015, Liu2018}. Instead, we perform scale-fusion using a normalized convolution layer which takes into account the confidence information embedded in different scales. We evaluate both approaches in our multi-scale architecture and our confidence-based approach NConv-HMS significantly outperforms the standard fusion approach NConv-SF-STD as shown in Table \ref{tab:1}. This clearly demonstrates the significance of utilizing confidence information for selecting the most confident data within the network.

\noindent\textbf{Comparison on the test set}: 
Here, we evaluate on the test set, which can only be performed on the benchmark server. Table \ref{tab:2} shows the error metrics for state-of-the-art methods published in the literature that are based on deep learning. SparseConv \cite{Uhrig2017} performs significantly better on the test set than the validation set, while DCCS-3-Layers maintains its performance. NN+CNN corresponds to performing nearest-neighbor filling for missing pixels and then train a CNN with the same architecture as \cite{Uhrig2017} to enhance the output. Our approach outperforms all published state-of-the-art methods on the test set. Contrary to the validation set, our approach outperforms DCCS-3-Layers on the test set.

\subsection{Qualitative Analysis}

\begin{table}[t]
\begin{center}
\begin{tabular}{R{2cm} | C{2.5cm} C{2.5cm} C{2.5cm} C{2.5cm}}
& SparseConv \cite{Uhrig2017} & NN+CNN \cite{Uhrig2017} & DCCS-3-Layers \cite{Chodosh2018} & NConv-HMS (Ours) \\
\hline 
MAE [m] & 0.48 & 0.41 & 0.44 & \textbf{0.37} \\
RMSE [m] & 1.60 & 1.41 & 1.32 & \textbf{1.29} \\
\hline 
\end{tabular}
\end{center}
\caption{Quantitative results on the test set. All the results are taken from the online KITTI depth benchmark \cite{Uhrig2017}. Our method outperforms all published methods 
on the benchmark.}
\label{tab:2}
\end{table}

\setlength{\belowcaptionskip}{-5mm}
\setlength{\tabcolsep}{1pt}
\begin{figure}[t]
\begin{tabular}{cc}
\includegraphics[width=0.48\textwidth]{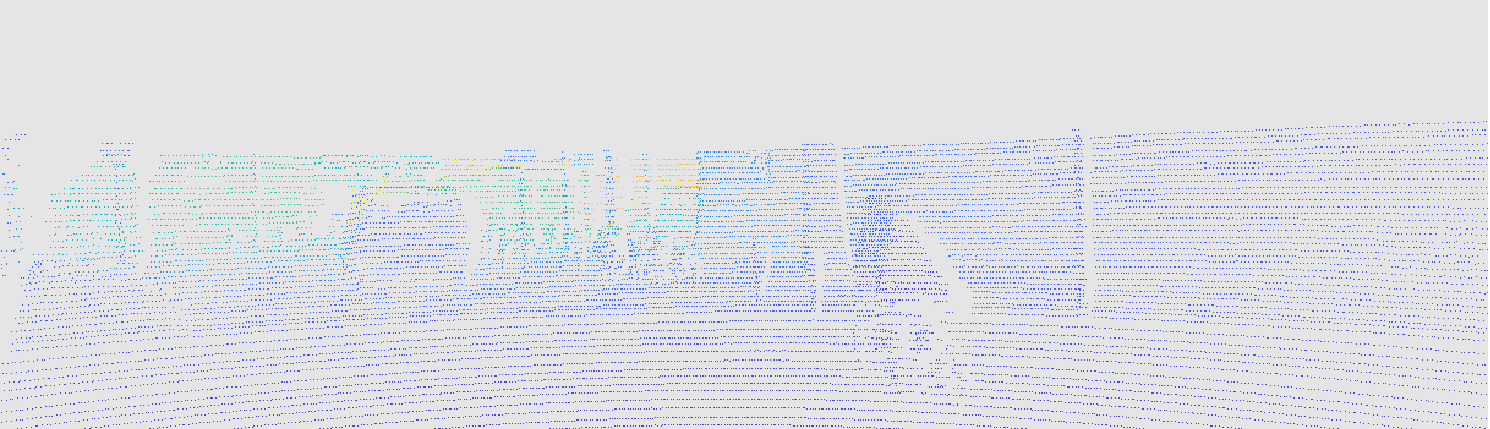}&
\includegraphics[width=0.48\textwidth]{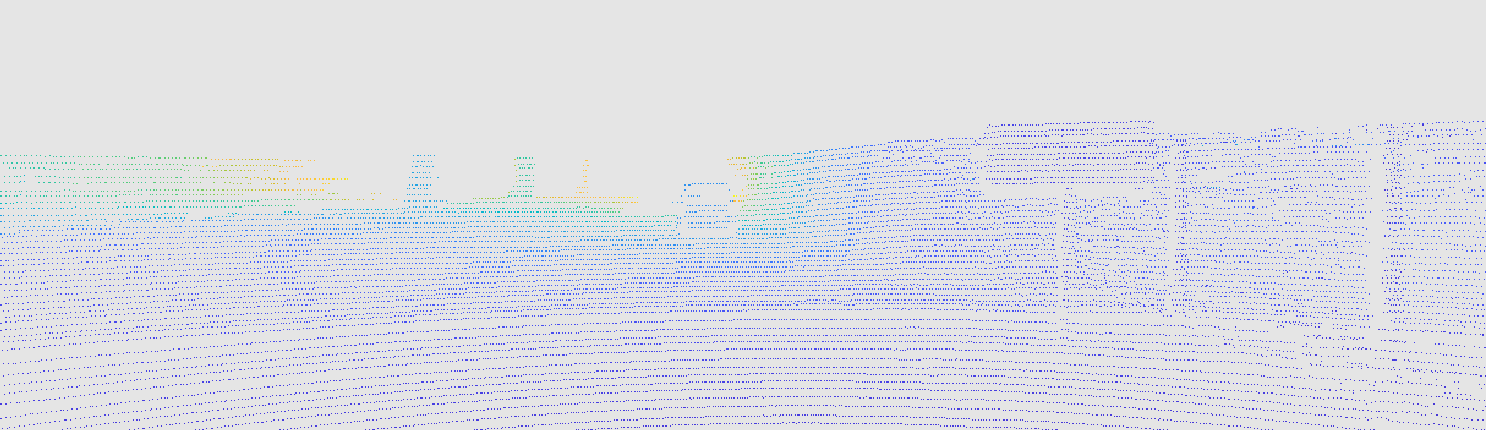} \\
\includegraphics[width=0.48\textwidth]{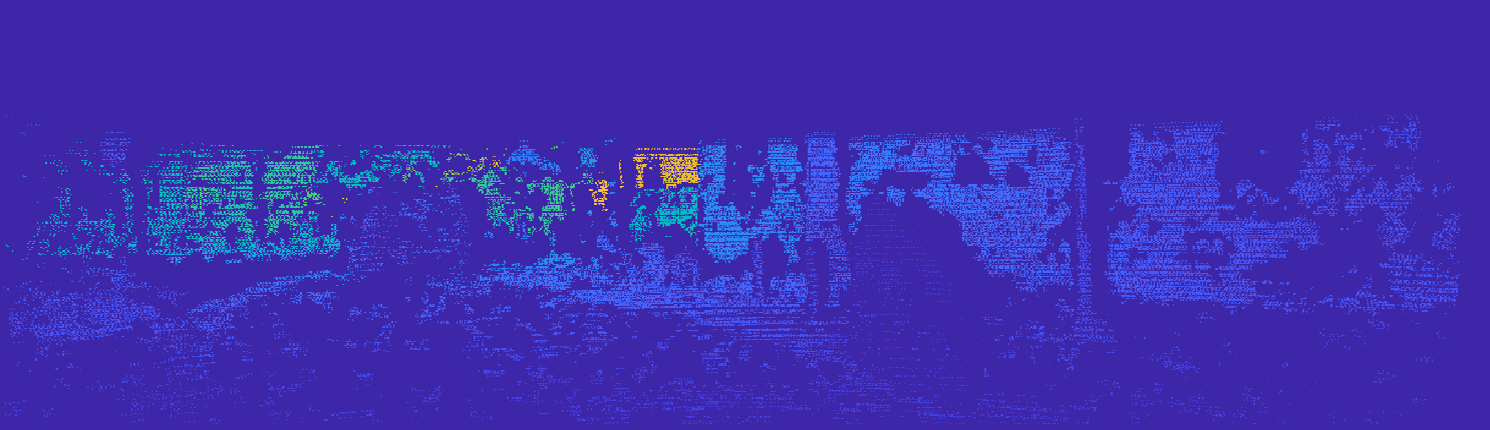}&
\includegraphics[width=0.48\textwidth]{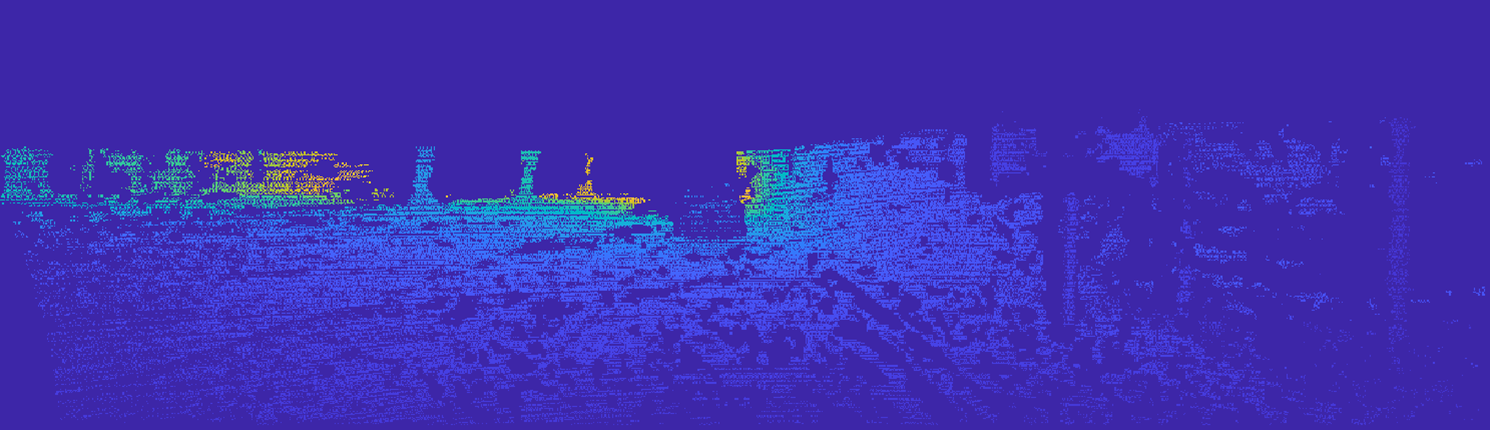} \\
\includegraphics[width=0.48\textwidth]{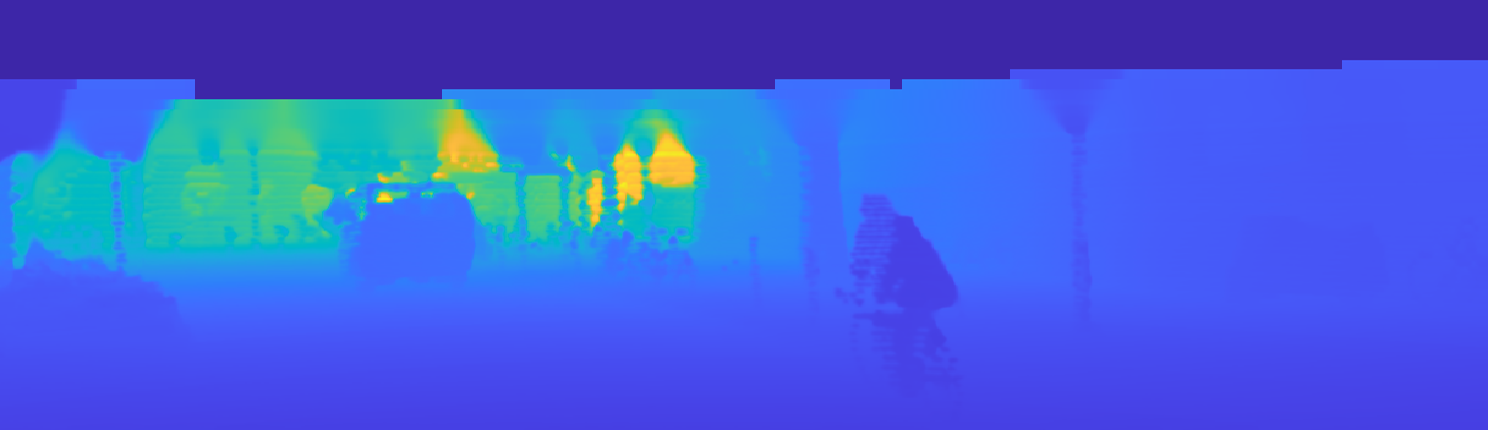}&
\includegraphics[width=0.48\textwidth]{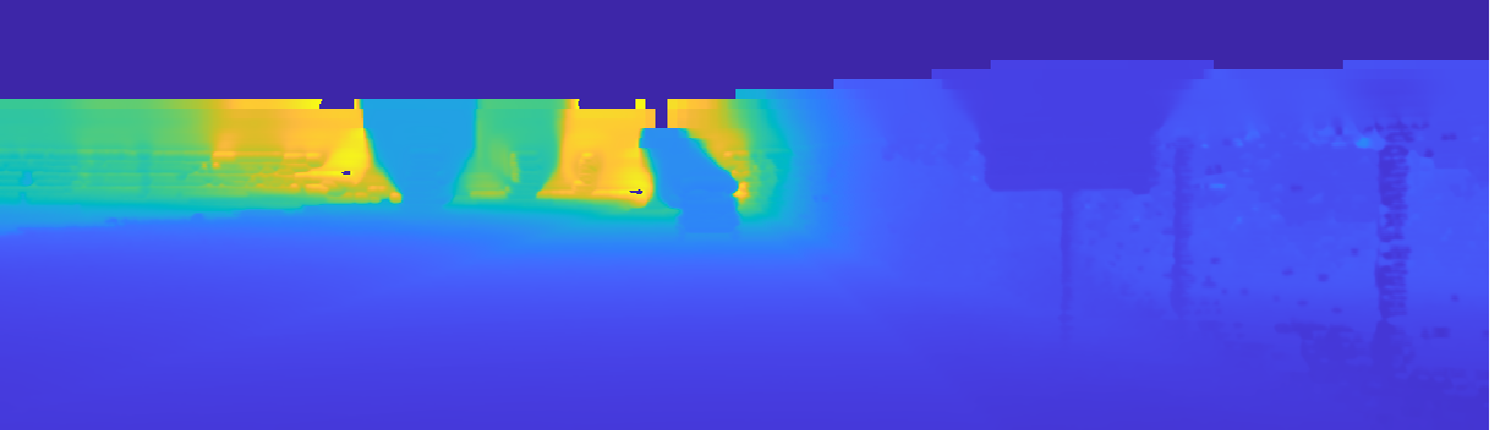} \\
\includegraphics[width=0.48\textwidth]{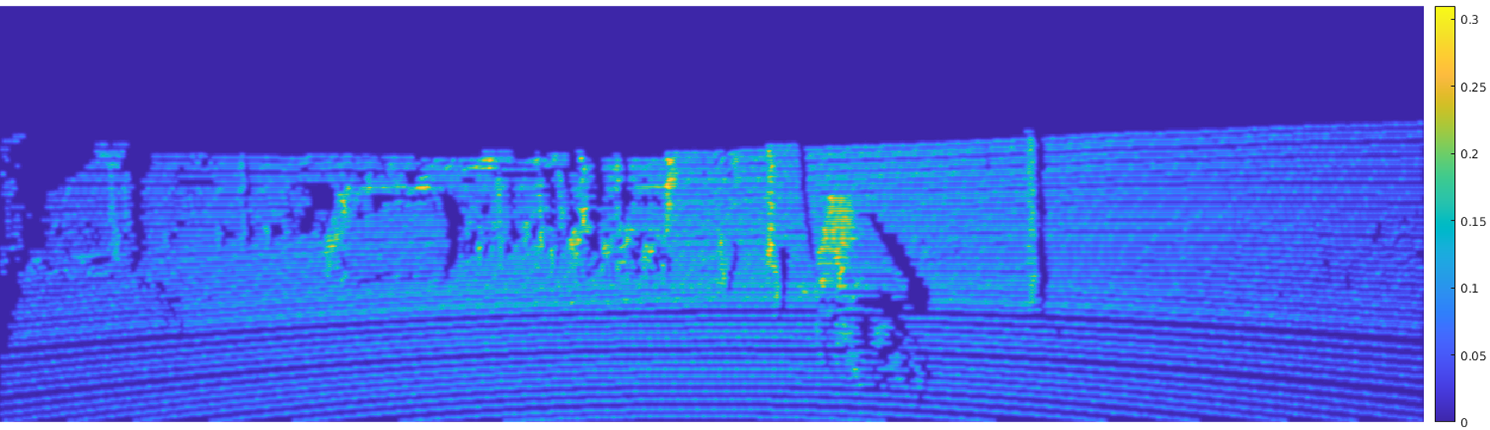}&
\includegraphics[width=0.48\textwidth]{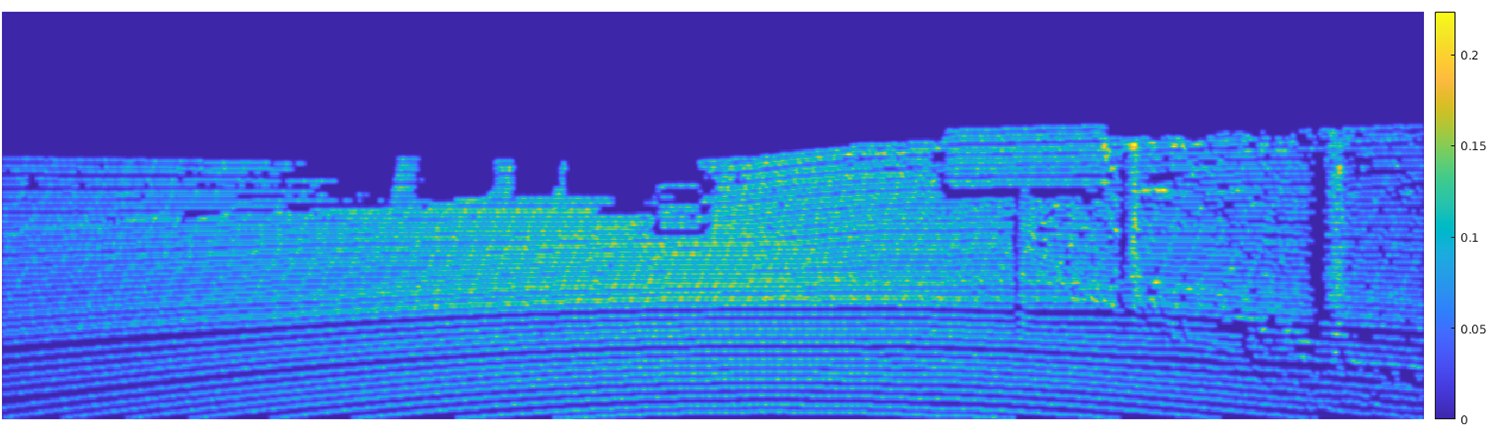} \\
\end{tabular}
\caption{Examples of scene depth completion using our mutli-scale architecture on the KITTI depth benchmark \cite{Uhrig2017}. \emph{First} row are the sparse projected LiDAR point clouds, \emph{second} row are the ground truth images, \emph{third} row are the dense outputs from our method, and the \emph{last} row are the output confidence maps. Our method performs favorably on densifying the sparse input, while providing a confidence map that indicates the output reliability.}
\label{fig:qual}
\end{figure}

To further analyze the impact of the proposed contributions, we perform a qualitative study on the KITTI depth benchmark \cite{Uhrig2017}. Figure \ref{fig:qual} shows examples for performing depth scene completion on two images from the benchmark. The input are projected LiDAR point clouds that are highly sparse. The ground truth images are not completely dense due to the strict outlier filtering adopted by \cite{Uhrig2017}. These missing data impose a big challenge on methods to learn a good representation. As shown in the figure, our multi-scale architecture performs very well on densifying the sparse input. Moreover, the output confidences from our method provide indication about how reliable the output depth maps are. At locations where neither input points nor groundtruth information is available, e.g. behind the cyclists or below the billboard, the output confidence is very low.  Further, the results show that regions in the center of the scene tend to have high confidence due to the high point cloud density in the input. This demonstrates that our method for confidence propagation enables the network to learn the prominence of different regions with respect to the groundtruth.

\begin{figure}[t]
\begin{tabular}{cc}
\includegraphics[width=0.48\textwidth]{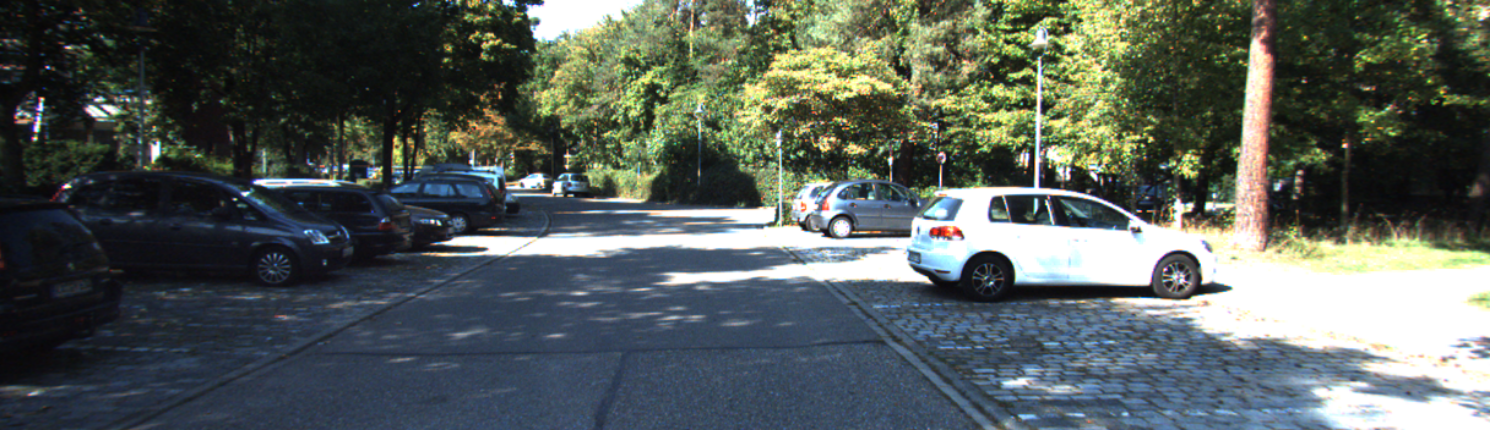}&
\includegraphics[width=0.48\textwidth]{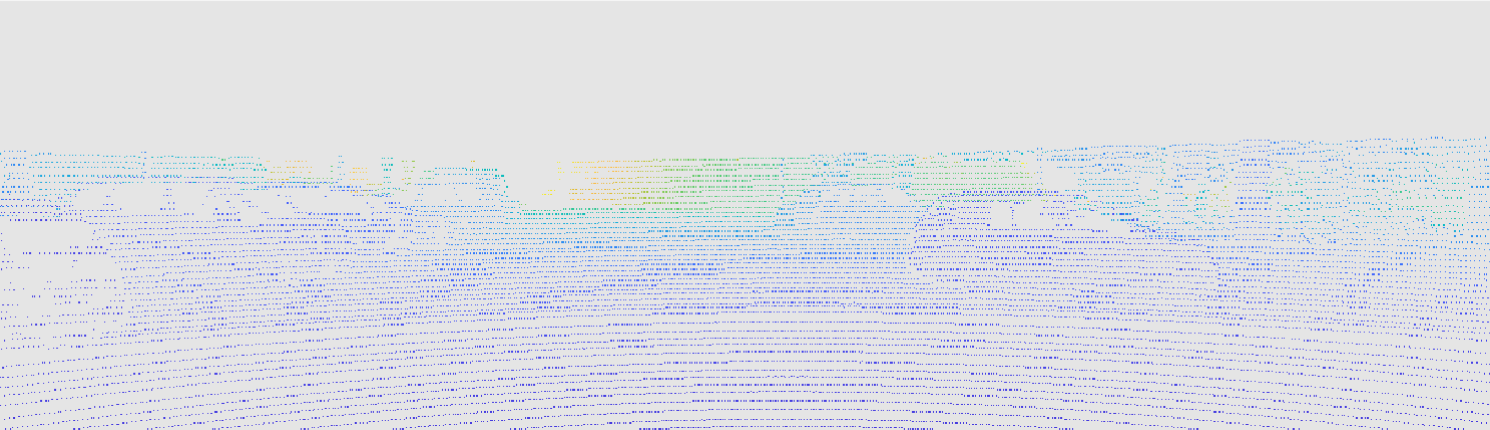} \\
\includegraphics[width=0.48\textwidth]{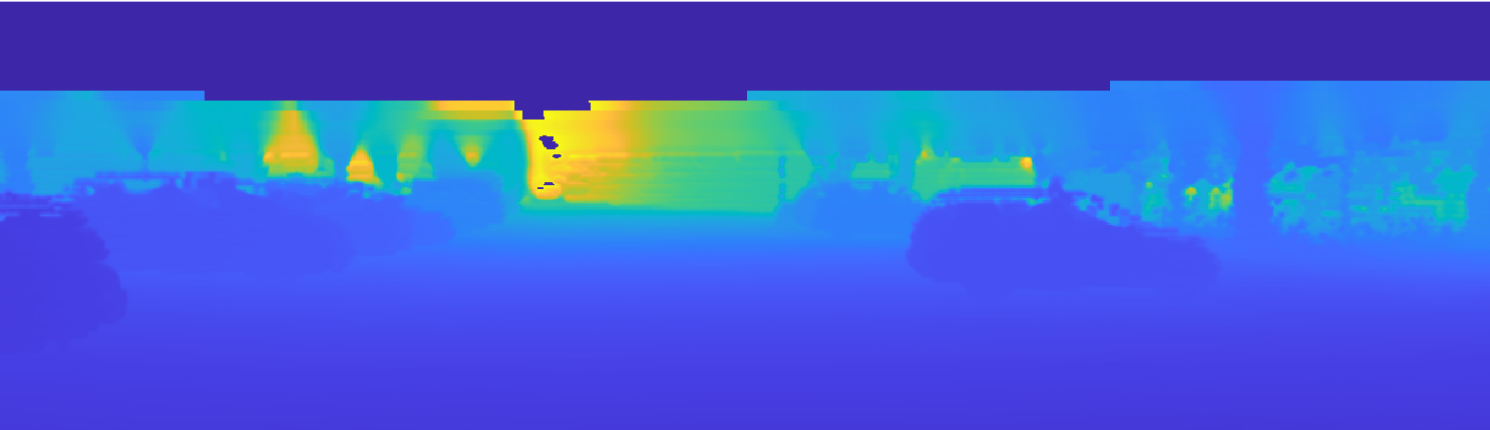}&
\includegraphics[width=0.48\textwidth]{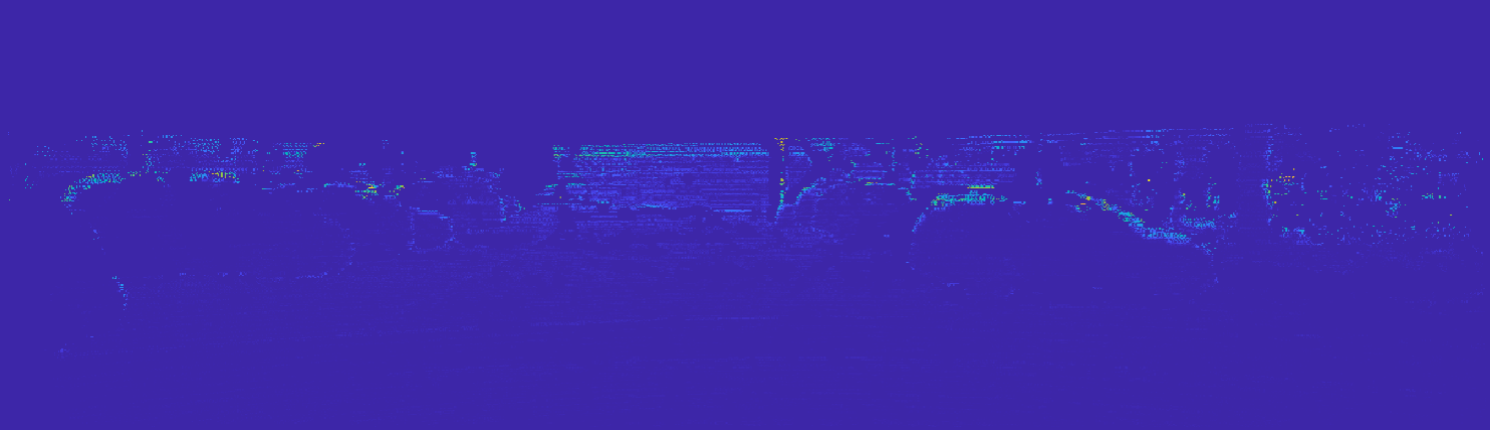} \\
\end{tabular}
\caption{An example of error analysis for our proposed method on KITTI Depth benchmark \cite{Uhrig2017}. \emph{Top-left} is the input RGB image, \emph{top-right} is the projected LiDAR point cloud,\emph{ bottom-left} is the output from our method and \emph{bottom-right} is the error map in logarithmic scale. The error is mainly distributed along edges and close to the horizon.}
\label{fig:err}
\end{figure}

\noindent\textbf{Error analysis}:
As discussed earlier, our single-scale architecture suffers from a limited receptive field and fails to predict values for regions above the horizon in some images. This leads to a significant increase in the RMSE. We addressed this problem by adopting a multi-scale architecture to cover the whole receptive field. This allows our method to perform well on the whole validation set. For the case of the multi-scale architecture, the error is mainly distributed along sharp edges and upon the horizon. This is likely due to the absence of structural information that could be found in RGB images. Figure \ref{fig:err} shows an example of where the largest errors of our method are located. Obviously, those errors are distributed along the vehicles edges and close to the horizon. This problem could be addressed by incorporating prior knowledge about the structure of the scene from the RGB image.

\setlength{\tabcolsep}{1pt}

%% file: sec/concl.tex
In this paper, we proposed an algebraically-constrained convolution layer for CNNs to tackle the issue of sparse and irregularly spaced input data. Unlike previous works, we treated the input masks as continuous confidences instead of binary
values and equip the sparse input with confidences. We further derived novel methods for determining the confidence from the convolution operation and propagating it to consecutive layers. A non-negativity constraints on the network weights is imposed to maintain the confidences within a valid range. Moreover, we introduced an objective function that simultaneously minimizes the data error while maximizing the output confidence. Comprehensive experiments are performed on the KITTI depth benchmark for scene depth completion. The results show that our approach achieves superior performance while requiring significantly fewer parameters. Finally, the continuous pixel-wise confidence map produced by our approach is shown to produce reasonable results enabling proper information fusion, state inference, and decision support.